\documentclass{article}

\usepackage{microtype}
\usepackage{graphicx}
\usepackage{subfigure}
\usepackage{booktabs} 
\usepackage{amsmath}
\usepackage{textcomp}

\usepackage{hyperref}

\usepackage[accepted]{icml2020}

\icmltitlerunning{A Federated F-score Based Ensemble Model for Automatic Rule Extraction}

\begin{document}

\twocolumn[
\icmltitle{
A Federated F-score Based Ensemble Model \\for Automatic Rule Extraction
}

\icmlsetsymbol{equal}{*}

\begin{icmlauthorlist}
\icmlauthor{Kun Li}{to}
\icmlauthor{Fanglan Zheng}{to}
\icmlauthor{Jiang Tian}{to}
\icmlauthor{Xiaojia Xiang}{to}
\end{icmlauthorlist}

\icmlaffiliation{to}{Everbright Technology Co. LTD, Beijing, China}

\icmlcorrespondingauthor{Kun Li}{likun3@ebchinatech.com}
\icmlcorrespondingauthor{Xiaojia Xiang}{xiangxiaojia@ebchinatech.com}


\vskip 0.3in
]

\printAffiliationsAndNotice{}  

\begin{abstract}
In this manuscript, we propose a federated F-score based ensemble tree model for automatic rule extraction, namely Fed-FEARE. 
Under the premise of data privacy protection, Fed-FEARE enables multiple
agencies to jointly extract set of rules both vertically and horizontally. 
Compared with that without federated learning, 
measures in evaluating model performance 
are highly improved. At present, Fed-FEARE has already been applied
to multiple business, including anti-fraud and precision marketing, in 
a China nation-wide financial holdings group.
\end{abstract}

\section{Introduction}
With the rapid integration of internet technology and traditional finance, 
more and more financial transactions and activities, such as third-party payment 
and online lending, have been digitalized. Taking online payment as an example, 
it contributed over 2.8 billion users with a value of almost 3.1 trillion  
US dollars worldwide in 2018. 
In companion with it, financial frauds trend to be more subtle and diversified.
According to Nilson report \cite{nilson2019fraud}, fraudulent activities cost about 11.2 billion US dollars worldwide in 2012, and the number has increased almost by $150\%$ up to 27.85 billion US dollars in 2018.

Currently, financial agencies defend against fraud attacks
by implementing decision-making engine using expert-defining rules,
which are always based on learning from expert experience and analyzing 
the existing frauds. It has been widely used in those above financial
scenarios and has achieved good results. In practice, however, 
this expert-defining rule system always suffers two fundamental issues: 
(i) it's difficult to learn effective rules due to the lack of fraud samples; 
(ii) owing to the delayed characteristic of fraud detection, it inevitably
suffers from not being updated in time, as well as high false alarm rate 
and expensive maintenance costs. Refer to \cite{bolton2002statistical}.

Meanwhile, traditional machine learning models also have similar issues \cite{bhattacharyya2011data,ngai2011the}. To improve the predictive ability of the model, 
a federated learning framework has been recently proposed even
at the cost of mass of training time, due to relatively heavy
encryption and communication for privacy-preserving. It \cite{bonawitz2017practical,yang2019federated} is an emerging frontier field studying privacy-preserving collaborative machine learning while leaving data instances at their providers locally. Federated learning enables multiple agencies to collaboratively learn a shared model while keeping all the training data stored on their own private database.

In order to take advantage of both rule system and federated learning, we propose a F-score based ensemble model for automatic rule extraction (FEARE) and implement it
in federated learning framework (Fed-FEARE). 
In the process of building a tree in FEARE, 
we employ maximizing F-score as loss function or partition criterion 
in each node in recursive manner. 
Combination of the multiple partition logic in child nodes then forms a rule 
from the built tree. Next, the data set covered by the rule is removed.
Repeating the above tree-building process on the remaining data, 
it finally results in the formation a set of rules from ensemble trees.
It should be noted that the rule-extraction method of FEARE 
is quit different from that of traditional decision tree \cite{Breiman:1984jka,quinlan1993c}.
For comparison, there are three major differences:
(1) loss function or partition criterion, 
F-score $vs$ Gini index or gain ratio; (2) learning a set of rules progressively $vs$ simultaneously; (3) every rule extracted from one tree with high quality.

With Fed-FEARE, we train an ensemble model and evaluate its performance 
on large scale real-world data sets from both 
a nation-wide joint-stock commercial bank (\textbf{BANK})
and
a cloud payment platform (\textbf{CLOUD PAY})
, two separated legal entities in a China nation-wide financial holdings group
with certain number of fraud cases. The experimental results show that recall
is greatly improved with high precision compared to that without Fed-FEARE.
Also, we apply horizontal Fed-FEARE to precision marketing. 
As expected, both precision and lift gain have obvious improvement.

The rest of this manuscript is organized as follows. In Sect. \ref{sec:Alg}, 
related work about F-score based ensemble model for automatic rule set extraction is proposed. Furthermore, the vertical and horizontal federated learning framework for the model is specifically discussed. Sect. \ref{sec:er} and \ref{sec:fm} give the details of our experimental 
results in both vertical and horizontal Fed-FEARE. 
Conclusions are presented in Sect. \ref{sec:con}.

\section{Algorithm for F-score based Ensemble Model for Automatic Rule Extraction and Its
Federated Learning Framework}\label{sec:Alg}

This section consists of two parts. Firstly, we propose a F-score based ensemble tree model for automatic risk-rule extraction. Secondly, we implement it in both the vertical and horizontal federated learning framework, including encryption and communication for privacy-preserving. 

\subsection{Loss Function, Pruning and Automatic Rule Set Extraction}
In general, rule can be evaluated by its precision and recall.
For an given class-labeled data set $D$ with $n_{target}$ target samples, 
$n_{cover}$ and $n_{correct}$ are data size for rule coverage 
and correct classification, respectively. Precision and recall
are thus defined as,

\begin{equation}\label{Pre}
    \begin{split}
    precision &=\frac{n_{correct}}{n_{cover}},\\
    recall &= \frac{n_{correct}}{n_{target}}.
    \end{split}
\end{equation}

Ideally, we hope to increase recall as much as possible 
under the condition of high precision. 
In the anti-fraud scenario, however, it's often unreliable to simply use precision or recall as rule measurement. High precision and high recall are always difficult to coexist.
For instance, rule one correctly classifies 80 of the 100 samples it covers, 
while the two samples covered by rule two are correctly classified. Though rule two 
has obviously higher precision, it is still not a better one due to its small coverage.
Likewise, coverage cannot be used as a measure of rule.
Therefore, it is necessary to construct other measures to evaluate rule. 

F-score ($\mbox{F}_{\beta}\mbox{-score}$), a weighted average of precision and recall,
can be treated as a rule measurement. Thus, it acts as an attribute (feature) splitting criteria. 
\begin{equation}\label{F-score}
    \resizebox{.7\linewidth}{!}{$
    \displaystyle
    \mbox{F}_{\beta}\mbox{-score}= (1 + \beta^2)\frac{precision \cdot recall}{\beta^2\cdot precision +  recall}.
$}
\end{equation}

$\beta = 1$, that is, precision and recall have the same weight. 
When the importance of precision is higher than recall, 
$\beta < 1$ can be set, and vice versa.
The F-score gain is defined as the difference before and after 
splitting an attribute of the data set.
Accordingly, an attribute with the highest F-score gain 
is chosen as the best splitting attribute of child node.
In order to achieve it, we need to calculate and find the best splitting point in an attribute. 
An attribute is sorted in descending order by value when it is numerical. 
For categorical attributes, some encoding methods are adopted to convert to numerical type.   
The average of each pair of adjacent values in an attribute with $n$ value, forms $n-1$ splitting points or values. As for this attribute, the point of the highest F-score gain can be seen as the best partition one.   
Furthermore, the best splitting attribute with the highest F-score gain can be achieved by traversing all attributes. As for the best splitting attribute, the instance space would be divided into two
sub-spaces at the best splitting point. Note that the top–down, recursive partition
will continue unless there is no attribute
that explains the target with statistical significance.

\begin{algorithm}[htbp]
\caption{Learning A "IF-THEN" Classification Rule of A Single Tree
}
\label{single:algorithm}
\textbf{Input}: D, the given class-labeled data set;\\
\textbf{Parameter}: max$\_$depth, $\beta$, pruning$\_$min\\
\textbf{Output}: a "IF-THEN" classification rule

\begin{algorithmic}[1] 
\STATE \textbf{Set} Rule$\_$Single = [], Max$\_$F-score = 0.0
\STATE \textbf{Set} Add\_Rule = \textbf{True}
\WHILE{depth $\leq$ max$\_$depth \textbf{and} Add\_Rule}
\STATE \textbf{Set}  Keep = \{ \}, Best$\_$Split = \{ \}
\STATE depth $\leftarrow $ depth + 1
\STATE Add\_Rule = \textbf{False}
\FOR{feature in features}
\STATE Keep[feature] = F-score$\_$Cal(D, feature, $\beta$) 
\ENDFOR
\FOR{feature in Keep}
\IF {feature's best F-score \textgreater Max\_F-score + pruning\_min}
\STATE Max$\_$F-score = feature's best F-score
\STATE \textbf{Add} Keep[feature] \textbf{to} Best$\_$Split
\STATE Add\_Rule = \textbf{True}
\ELSE
\STATE continue
\ENDIF
\ENDFOR
\STATE \textbf{Add} Best$\_$Split \textbf{to} Rule$\_$single
\STATE D $\leftarrow $ D $\setminus$ \{
Samples covered by Rule$\_$single\} 
\ENDWHILE
\STATE \textbf{return} Rule$\_$single
\end{algorithmic} 
\end{algorithm}

In the tree-building process, due to noise and outliers in the data set, many branches merely represent these abnormal points, resulting in model overfitting. 
Pruning can often effectively deal with this problem. That is, using statistics to cut off unreliable branches. Since none of the pruning methods is essentially better than others, we use a relatively simple pre-pruning. That is, if the F-score gain was less the threshold, node partition 
would stop. Thus, a smaller and simpler tree is constructed after pruning. 
Naturally, decision-makers prefer less complex rules, since they may be considered 
more comprehensible and robust in business perspective.

Algorithm \ref{single:algorithm} presents a typical algorithmic framework for top–down inducing of a rule tree using growing and pruning. It uses a greedy depth-first strategy
in constructing tree in a recursive manner. In each iteration, 
the algorithm considers the splitting of the training data set using F-score gain as partition criteria. 
It removes those instances that is not covered by the node logic in the original data set. 
As a result, each child node hierarchically subdivides the training data set into smaller subsets, until stopping criteria is satisfied.

As we can see from Algorithm \ref{single:algorithm}, F-score$\_$Cal is a function to calculate the F-score before and after the child node partition and find one attribute (feature)'s best splitting. $max\_depth$ and $pruning\_min$ are the depth of the tree and threshold of F-score gain, respectively. 
$\beta$ is the parameter in Eq.\ref{F-score}, where $\beta=1$ corresponds to 
F$_{1}$-score. $Keep$ records the F-score based partition gain and calculation logic symbol in all attributes. And Rule$\_$Single is a rule formed by a single tree. Tracing the path from the root to child node in the tree, a "IF-THEN" classification rule is thus extracted. 

\begin{algorithm}[htbp]
\caption{Learning A Set of "IF-THEN" Classification Rules}
\label{multi:algorithm}
\textbf{Input}: D, the given class-labeled data set;\\
\textbf{Parameter}: tree$\_$number, max$\_$depth, $\beta$\\
\textbf{Output}: a set of "IF-THEN" classification rules

\begin{algorithmic}[1] 
\STATE \textbf{Set} Rule$\_$Set = \{\}, number = 0
\WHILE{number $\leq$ tree$\_$number}
\STATE rule = Single$\_$Risk$\_$Rule(D,max$\_$depth,$\beta$)\\
\STATE \textbf{Add} rule \textbf{to} Rule$\_$Set  
\STATE D $\leftarrow $ D $\setminus$ data set covered by rule 
\STATE number $\leftarrow$ number + 1
\ENDWHILE
\STATE \textbf{return} Rule$\_$Set
\end{algorithmic} 
\end{algorithm}

Next, the censored data sets constitute the remaining data set. 
Repeating the above tree-building process on the latter, 
a set of rules are automatically extracted from the built trees.
As shown in Algorithm \ref{multi:algorithm},  Rule$\_$set contains and 
returns a set of rules. 

\subsection{FEARE in Federated Learning Framework}
Federated Learning firstly focuses on the horizontal structure, in which each node has a subset of 
data instances with complete data attributes. There are also many researches studying the vertical 
federated learning structure where the data set is vertically partitioned and owned by different data providers. 
Each data provider holds a disjoint subset of attributes for all data instances.
For both horizontal and vertical federated learning, the target is to learn a machine learning model 
collaboratively without transferring any data from one data provider to another. In our security definition, 
all parties are honest-but-curious.

\begin{figure}[htbp]
\centering
\includegraphics[width=0.95\linewidth]{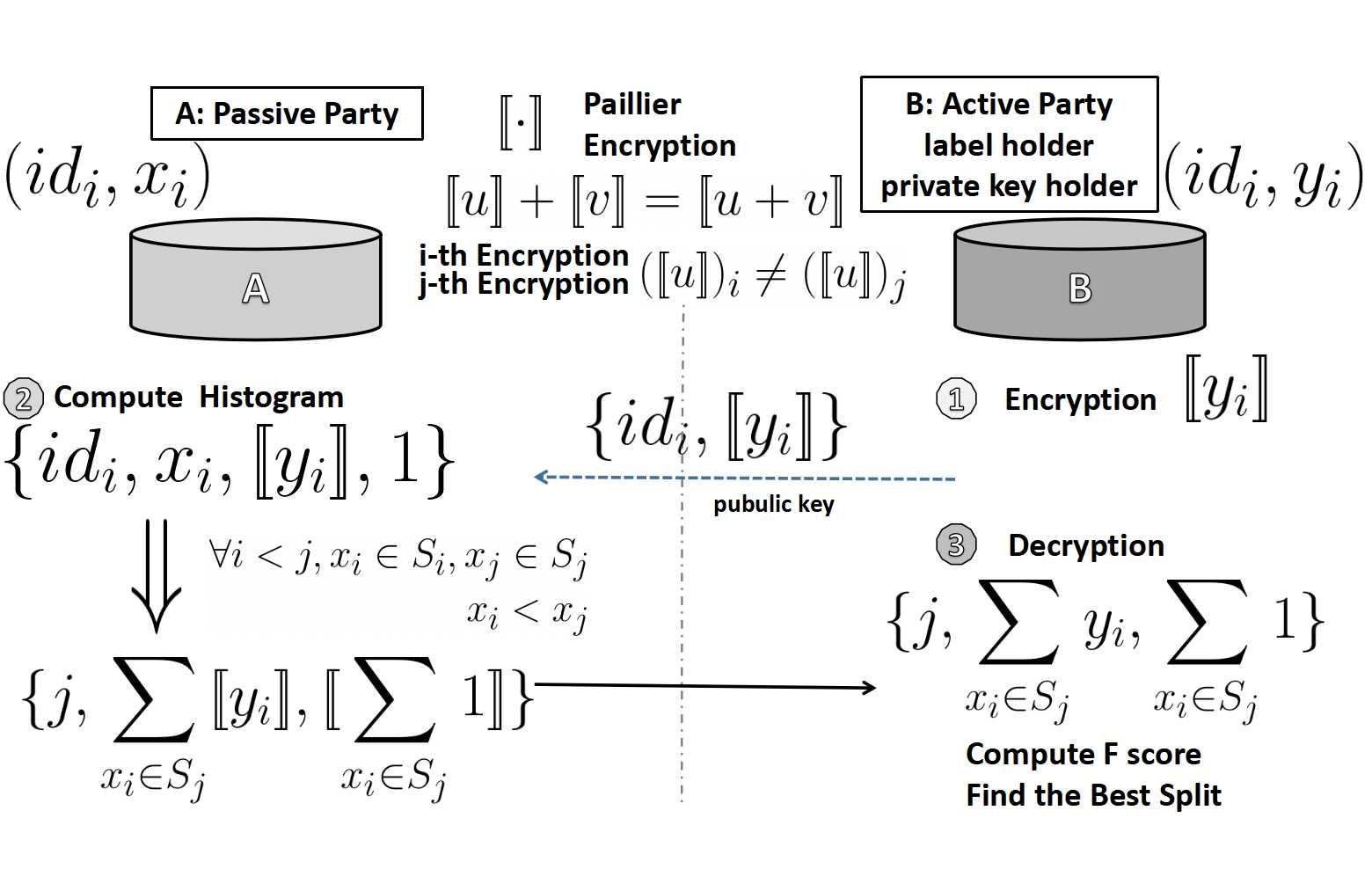} 
\caption{ A Vertical Federated Learning Framework of F-score Calculation and the Best Split Search of one feature from Passive Party.}
\label{fig:VFL} 
\end{figure}

The main challenge in Fed-FEARE is how to calculate the F-score. The key tools to this challenge 
are partially homomorphic encryption schemes.
Paillier encryption \cite{Paillier1999}  allows any party can encrypt their
data with a public key, while the private key for decryption
is owned by the third party. With this additively
homomorphic encryption we can compute the additive
of two encrypted numbers as well as the product of an
unencrypted number and an encrypted one, which can be denoted
as $[\![u]\!] + [\![v]\!] = [\![u + v]\!]$, $v[\![u]\!] = [\![vu]\!]$ by using $[\![\cdot]\!]$ as the encryption operation. Moreover, another advantage of Paillier encryption is that the results of each encryption of $u$ are different. Therefore, the encrypted labels $[\![y_i]\!]$, where $y_i \in \{0,1\}$, will not lead to information leakage.

\begin{figure*}[htbp]
\centering
\includegraphics[width=0.62\textwidth]{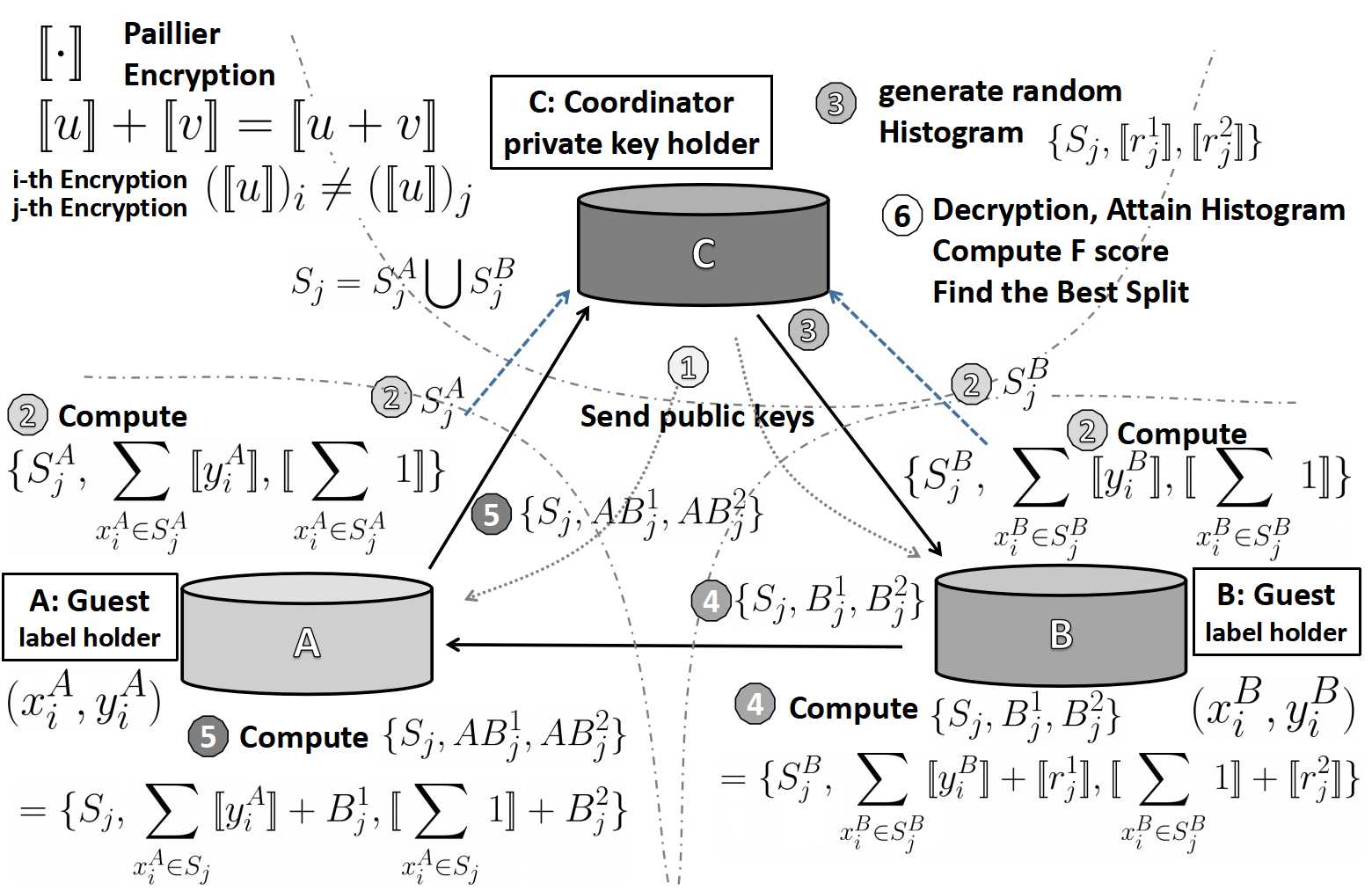} 
\caption{ A Horizontal Federated Learning Framework of F-score Calculation and the Best Split Search of One Feature.}
\label{fig:HFL} 
\end{figure*}

In \textbf{Vertical Federated Learning}, we follow the notation in \cite{cheng2019secureboost:}. The data set is vertically
partitioned and distributed on two honest-but-curious
private \textbf{Party A} (the passive data provider with only features
) and \textbf{Party B} (the active data provider with
features and labels $y_i$). The F-score calculation of the features from \textbf{Party B} is the same issue as non-federated case. For the features from \textbf{Party A}, the F-score calculation is still feasible. With the help of Paillier encryption, the vertical federated learning framework for F-score calculation and FEARE can be designed as Figure \ref{fig:VFL}. Due to the labels belong to \textbf{Party B}, the rule-set is achieved by \textbf{Party B}. However the value information of the feature from \textbf{Party A} is encoded by $(j,S_j)$. The splitting point founded by \textbf{Party B} is in the form of $j$ instead of $S_j$. Therefore, this framework is secure for passive data provider, even in multi passive parties case.

In \textbf{Horizontal Federated Learning}, we follow the notation in \cite{yang2019Quasi}. The data set is horizontally
partitioned and distributed on at least two honest-but-curious
private \textbf{Party A} (the guest data provider with features and labels $y^A_i$
) and \textbf{Party B} (the guest data provider with
features and labels $y^B_i$). The F-score calculation of the features for the data from \textbf{Party A\&B} is more complex than the issue in vertical case. Because the value information of feature is shared by \textbf{Party A\&B}, the histogram of each party should not be shared with any other party. For the purpose of privacy-preserving, we designed the horizontal federated learning framework shown in  Figure \ref{fig:HFL}. An honest-but-curious third party, i.e. the \textbf{coordinator}, is introduced here. For Paillier encryption, the private key for decryption is owned by \textbf{coordinator} and any party can encrypt their data with a public key. After receiving the value information $S^A_i$ and $S^B_j$ from guest parties, the \textbf{coordinator} sends a encrypted random histogram of one feature to one party
(\textbf{Party B} in Figure \ref{fig:HFL}). The calculation of F-score  can be accomplished by \textbf{coordinator} (Step 6 in Figure \ref{fig:HFL}), when the encrypted histogram of one feature comes back. In this framework, the guest parties only know the histogram of one feature based on their own data. The \textbf{coordinator} knows the histogram of features based on the whole data. The final rule-set will be shared by all parties.

\section{Financial Anti-Fraud within Horizontal Fed-FEARE Framework}\label{sec:er}
This section is organized with three parts. Firstly, we show parameter assignment in training a model. Secondly, we introduce the data sets that are tested in our horizontal Fed-FEARE framework. Finally, we demonstrate the results of our experiments.

\subsection{Parameter Assignment}
There are only four parameters in Fed-FEARE: $max\_depth$, $tree\_{number}$, pruning threshold ($pruning\_min$) and weight factor $\beta$. Generally, two main factors of business logic and the difficulty in online deployment, determine parameter assignment. Due to the requirement of model generalization and its interpretability, $max\_depth$ is set as 3. That is, the business logic of a rule is always less than or equals to three.
And, $tree\_{number}$ is defined as 3, implying that 
a set of no more than three rules is used to solve specific anti-fraud problems. It can avoid insufficient coverage due to too few trees or low accuracy due to overmuch trees, as well as online maintenance and other issues. 
Pruning threshold ($pruning\_min$) is fixed as constant, 0.01. 
Moreover, $\beta = 1.0$ is often set, indicating that both recall and precision
are of equal importance. 
It can be adjusted according to the business target of pursuing high precision or high recall.

\subsection{Data set Description}
For risk management in the financial anti-fraud, variable names of the relevant data in finance agencies won't be disclosed. 

With the usage of horizontal Fed-FEARE, 
\textbf{BANK}
trains an ensemble tree model with
\textbf{CLOUD PAY},
which has China's convenience payment data. 
Note that the target variable is coded as 1 to indicate default (according to a default or fraud definition chosen by the bank) and 0 to indicate non-default. From the \textbf{BANK}, there are 75,295 non-defaults and only 20 defaults (events or frauds) in the data set, separately. 
Combining with 60 defaults from \textbf{CLOUD PAY} horizontally, 
a training data set of 75,375 observations is thus formed. 
It corresponds to a highly imbalanced ratio of positive and negative 
instances, about 1:940. 
The total data set consists of 25 variables from the training data set. 
Different from the traditional tree model, Fed-FEARE, maintains their models in both 
agencies, without any private data transferring.

\subsection{Results and Discussions}

\begin{figure*}[htbp]
\centering
\includegraphics[width=0.33\textwidth]{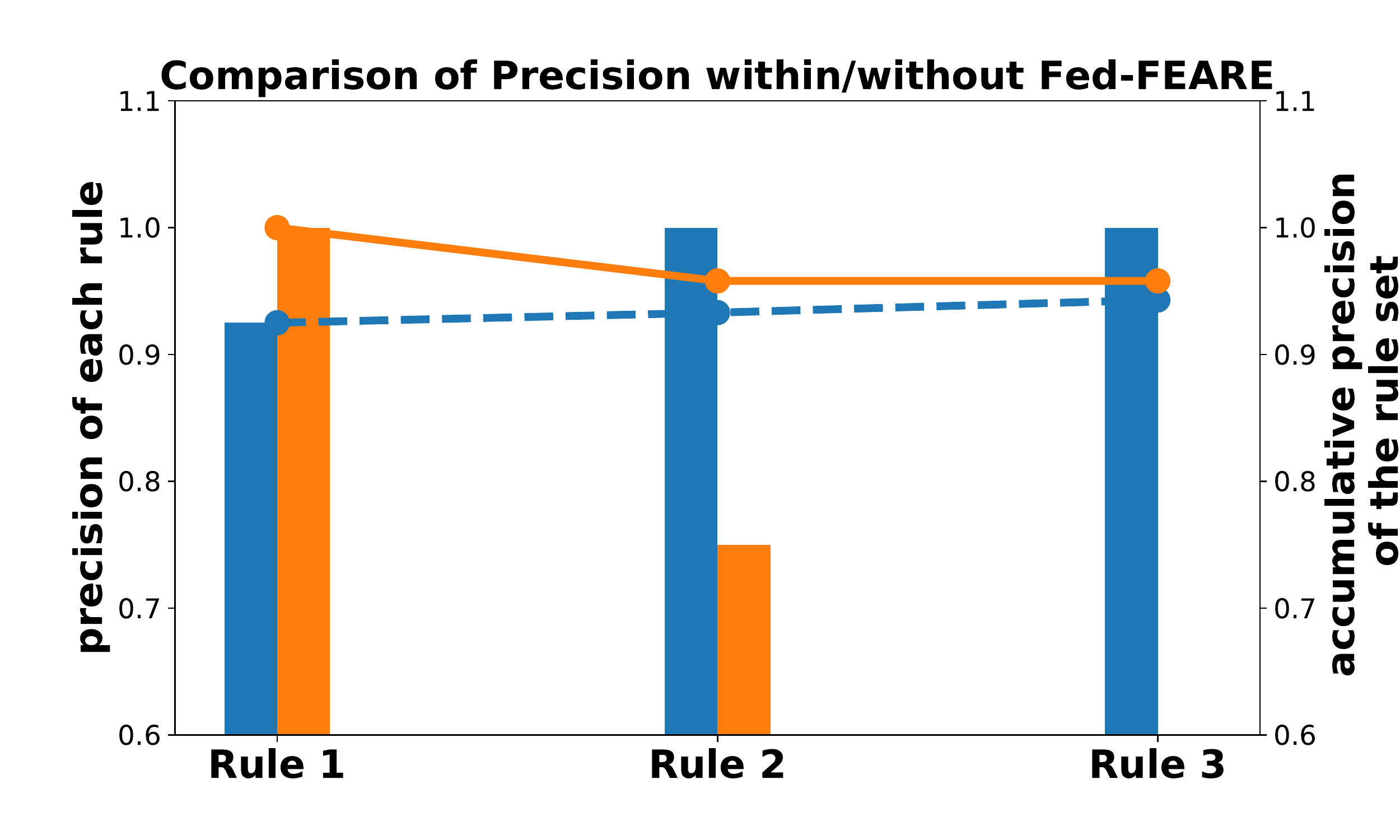} 
\includegraphics[width=0.33\linewidth]{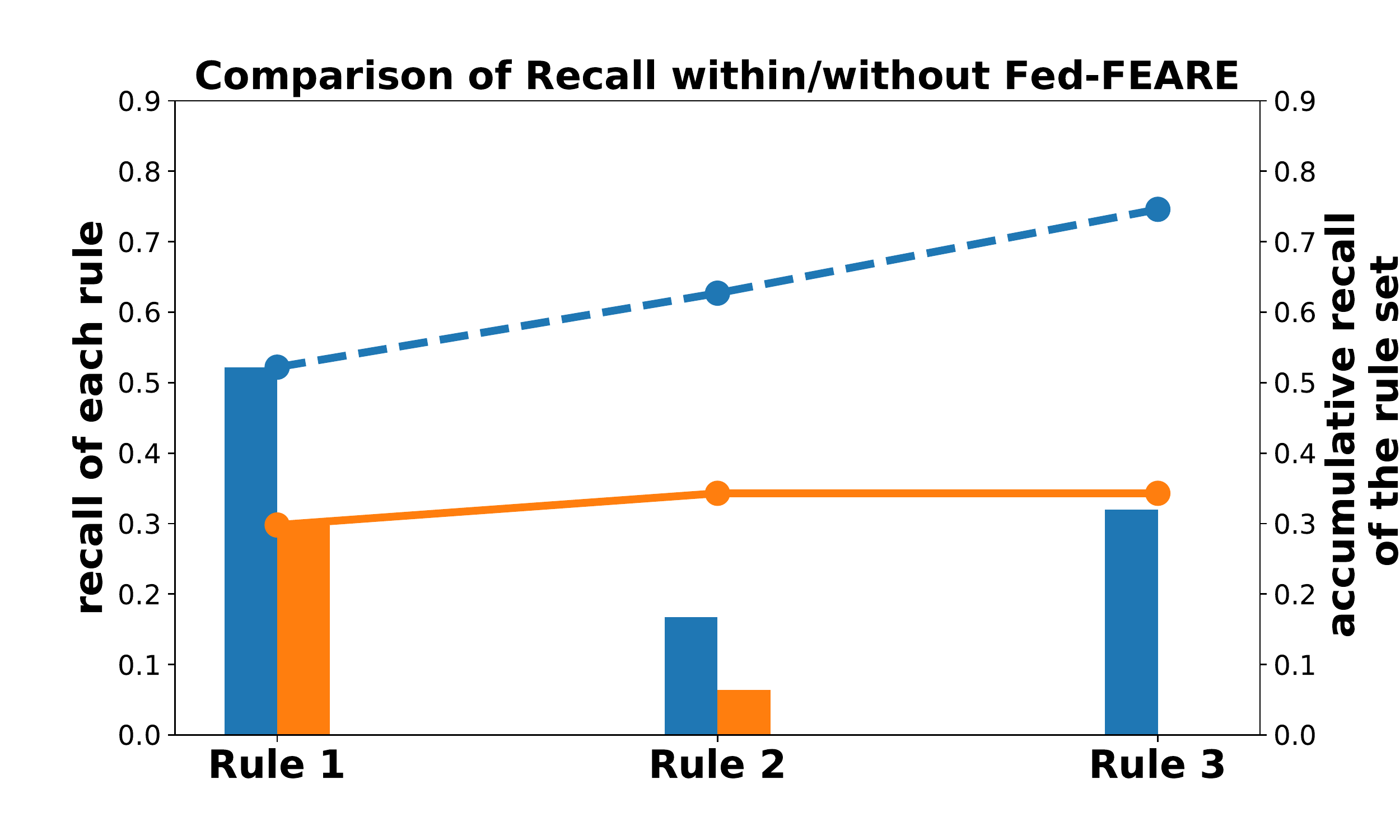} 
\includegraphics[width=0.33\linewidth]{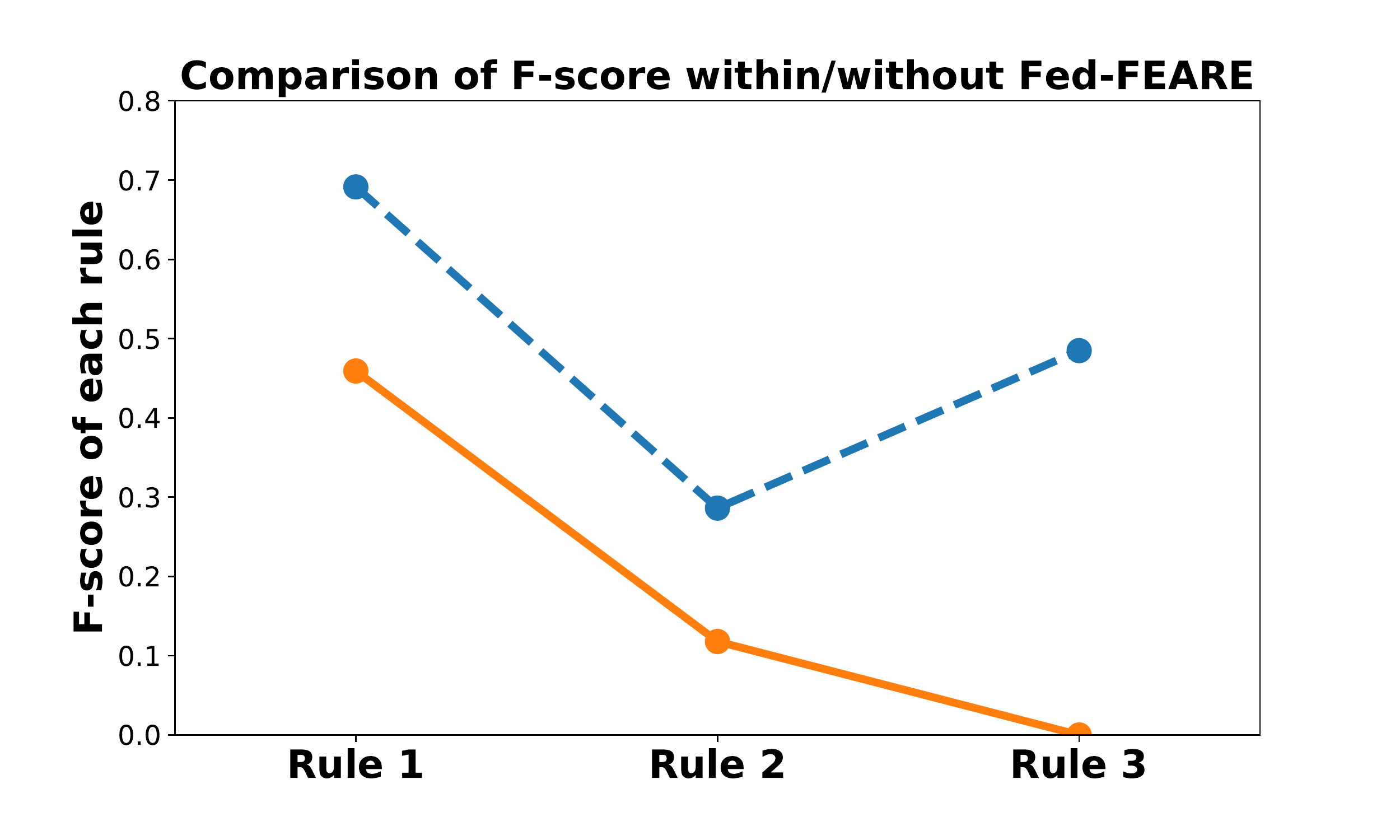} 
\caption{ Precision (left, bars), cp (left, lines), recall (middle, bars), cr (middle, lines) and F-score (right) within (blue) and without (orange) our horizontal Fed-FEARE.}
\label{fig:PR} 
\end{figure*}

Among these variables, 15 of them are characterized as identity traits, 
consumption grade, net assets, loan amount etc. The rest variables  
represent payment of apartment utilities (including electricity,
water, gas and property charges) and mobile communication
costs.

\begin{table}[htbp]
\centering
\resizebox{\linewidth}{!}{
\begin{tabular}{llll}
\hline
rule number  &"1"  &"2"  &"3"\\
\hline
node logic     &var$\_$12 $\leq$ -4.8  &var$\_$17 $\leq$ -3.0  &var$\_$14 $\leq$ -4.7 \\
node logic     &var$\_$1 $>$ -26.6    &var$\_$14 $>$ 1.3   &var$\_$10 $\leq$ -2.27\\
node logic     &null                  &var$\_$5 $>$ -3.3     &var$\_$2 $\leq$ 3.2                 \\
pi             &0.08$\%$              &0.011$\%$               &0.009$\%$           \\
cpi            &0.08$\%$              &0.091$\%$               &0.10$\%$           \\
F-score        &0.75                  &0.44                    &0.52             \\
precision      &83.0$\%$              &90.0$\%$                &88.8$\%$             \\
recall         &68.0$\%$              &29.0$\%$                &36.3$\%$       \\
cp             &83.0$\%$              &84.0$\%$                &84.6$\%$          \\
cr             &68.0$\%$              &72.5$\%$                &82.5$\%$           \\
\hline
\end{tabular}
}
\caption{Rule set and its corresponding proportion of instances (pi), cumulative proportion of instances (cpi), F-score, precision, recall, cumulative precision (cp) and cumulative recall (cr) within our horizontal Fed-FEARE framework (Financial Anti-Fraud).}
\label{tab:plain11}
\end{table}

\begin{table}[htbp]
\centering
\resizebox{\linewidth}{!}{
\begin{tabular}{llll}
\hline
rule number  &"1"  &"2"  &"3"\\
\hline
node logic     &var$\_$12 $\leq$ -5.6 &var$\_$14 $\leq$ -10.2  &var$\_$18 $>$ 3.4 \\
node logic     &var$\_$21 $>$ 30073   &var$\_$20 $\leq$ 0.99   &var$\_$1 $>$ 0.98\\ 
node logic     &var$\_$2 $>$ 2.25     &null                    &null                 \\
pi             &0.0172$\%$            &0.0027$\%$              &0.0013$\%$           \\
cpi            &0.0172$\%$            &0.0199$\%$              &0.021$\%$           \\
F-score        &0.72                  &0.4                     &0.28             \\
precision      &92.3$\%$              &100.0$\%$               &100.0$\%$             \\
recall         &60.0$\%$              &25.0$\%$                &16.6$\%$       \\
cp             &92.3$\%$              &93.3$\%$                &93.7$\%$          \\
cr             &60.0$\%$              &70.0$\%$                &75.0$\%$           \\
\hline
\end{tabular}
}
\caption{Rule set and its corresponding statistical indicators using only bank data (Financial Anti-Fraud).}
\label{tab:plain12}
\end{table}

According to Table \ref{tab:plain11} and Table \ref{tab:plain12}, 
it can be seen that rule sets have changed remarkably. 
We can see clearly that rule "1" and "2" evolves from var$\_$12 $\leq$ -5.6 $\&$ var$\_$21 $>$ 30073 $\&$ var$\_$2 $>$ 2.25 and var$\_$14 $\leq$ -10.2 $\&$  var$\_$20 $\leq$ 0.99
to var$\_$12 $\leq$ -4.8 $\&$ var$\_$1 $>$ -26.6 and
var$\_$17 $\leq$ -3.0 $\&$ var$\_$14 $>$ 1.3 $\&$ var$\_$5 $>$ -3.3, respectively. 
As well for rule "3", var$\_$18 $>$ 3.4 $\&$ var$\_$1 $>$ 0.98 becomes var$\_$14 $\leq$ -4.7 $\&$ var$\_$10 $\leq$ -2.27 $\&$ var$\_$2 $\leq$ 3.2.
It leads to significant changes in child node logic of both two rule sets in identifying fraud cases.
Within our horizontal Fed-FEARE framework, F-score in both rule "1" and "2" have an average increment of about 7$\%$, while F-score in rule "3" reaches 0.52 from 0.28 .
Also, there are evident changes in terms of accumulative precision (cp) and accumulative recall (cr). The former decreases from 93.7$\%$ to 84.6$\%$, while the latter increases from 75.0$\%$ to 82.5$\%$, separately.

Moreover, a new data set is used to verify the generalization of both two rule sets.
It consists of 10,000 non-defaults and 67 defaults in all.
We can see clearly from Figure \ref{fig:PR} that 
there is no data instance covered by rule "3" without federated learning.
As a result, cp trends to be equal in the above two rule sets.
And, F-score of each rule with federated learning are greatly improved due to data enrichment.
They increase from 0.46, 0.12 and 0.0 to 0.69, 0.29 and 0.48, respectively.  
Eventually, with approximately equal cp, cr in our horizontal Fed-FEARE framework reaches 74.6$\%$ from 34.3$\%$,  
with an evident increment of more than 117$\%$.

Therefore, based on horizontal Fed-FEARE, frauds and non-frauds will be classified more easily 
by our rule system with clear business explanation. 

\section{Precision Marketing within Vertical Fed-FEARE Framework}\label{sec:fm}

We further extend our algorithm framework to precision marketing for new customer activation.
Taking advantage of our algorithmic framework, 
\textbf{BANK}
trains a Fed-FEARE model with
\textbf{CLOUD PAY} vertically. 

\begin{table}[htbp]
\centering
\resizebox{\linewidth}{!}{
\begin{tabular}{llll}
\hline
rule number  &"1"  &"2"  &"3"\\
\hline
node logic     &var$\_$0 $>$ 0    &var$\_$7 $>$ 36.4  &var$\_$0 $>$ 0 \\
node logic     &var$\_$0 $\leq$ 1  &null                 &var$\_$9 $>$ 990\\
node logic     &null               &null                &null                 \\
pi             &5.6$\%$            &7.4$\%$             &2.1$\%$               \\
cpi            &5.6$\%$            &13$\%$              &15.1$\%$                 \\
F-score        &0.14               &0.07                &0.09             \\
precision      &4.7$\%$            &1.99$\%$            &3.7$\%$             \\
recall         &27.8$\%$           &21.2$\%$            &15.3$\%$       \\
cp             &4.7$\%$            &3.1$\%$             &3.2$\%$              \\
cr             &27.8$\%$           &43.1$\%$            &51.8$\%$                 \\
cl             &4.96               &3.3                 &3.4\\
\hline
\end{tabular}
}
\caption{Rule set and statistics indicators within our vertical Fed-FEARE framework (Precision Marketing).}
\label{tab:plain21}
\end{table}

There are 5,438,267 observations in the data set. Note that the target variable is coded as 1 and 0 to indicate activation and non-activation, respectively. There are 51,203 activation and 5,387,064 non-activation in the data set, corresponding to a ratio of positive and negative instances, about 1:105. The total data set consists of 10 variables from the above two agencies. 
In this business scenario, $\beta = 0.5$ is often set, indicating that precision are relatively more important than recall, while other parameters are the same as above.

\begin{table}[htbp]
\centering
\resizebox{\linewidth}{!}{
\begin{tabular}{llll}
\hline
rule number  &"1"  &"2"  &"3"\\
\hline
node logic     &var$\_$0 $>$ 0     &var$\_$0 $>$ 0       &var$\_$3 $<$ 24 \\
node logic     &var$\_$0 $\leq$ 1  &null                 &null\\
node logic     &null               &null                 &null                 \\
pi             &5.6$\%$            &18.2$\%$             &22.7$\%$               \\
cpi            &5.6$\%$            &23.8$\%$             &46.5$\%$                 \\
F-score        &0.14               &0.07                 &0.03             \\
precision      &4.7$\%$            &1.89$\%$             &0.7$\%$               \\
recall         &27.8$\%$           &48.4$\%$             &45.8$\%$       \\
cp             &4.7$\%$            &2.5$\%$              &1.6$\%$              \\
cr             &27.8$\%$           &48.5$\%$             &79.8$\%$                 \\
cl             &4.96               &2.64                 &1.7\\
\hline
\end{tabular}
}
\caption{Rule set and its corresponding statistical indicators using only bank data (Precision Marketing).}
\label{tab:plain22}
\end{table}

According to Table \ref{tab:plain21} and Table \ref{tab:plain22}, 
rule "1" in two cases is exactly the same, while rule "2" and rule "3" change remarkably. 
We can see that both of them evolves from var$\_$0 $>$ 0 
and var$\_$3 $<$ 24 to var$\_$7 $>$ 36.4 and var$\_$0 $>$ 0 $\&$ var$\_$9 $>$ 990, respectively. 
It leads to significant change in business logic. Moreover, 
it is obvious that the performance in terms of cp and cumulative lift (cl) is greatly improved due to data enrichment. 
Compared to the rule set
using only 
bank data, cp of vertical Fed-FEARE model has an increase of more than 100$\%$, reaching 3.2$\%$. 
Correspondingly, cl of 3.4 has an evident improvement with raise by 100$\%$. 
It means that more target customers can be transformed with less marketing resource.

As a result, we are able to 
do precision marketing to target customers identified by our rule system.
Next, we will cooperate with more agencies such as Insurance companies, Trust companies, travel agencies, 
and E-commerce platforms in multiple business scenarios with the help of our Fed-FEARE.

\section{Conclusion}\label{sec:con}
This manuscript proposes a F-score based ensemble tree model in
federated learning for automatic rule set extraction, 
Fed-FEARE in short. It is applicable to multiple business scenarios, 
including anti-fraud,precision marketing etc.
Compared with that without Fed-FEARE, 
measures in evaluating model performance 
are highly improved. 
Fed-FEARE not only has the characteristics of fast calculation and strong portability, but also ensures interpretability and robustness.

\section*{Acknowledgements}
The authors gratefully acknowledge Haiying Han, Yiming Cheng, Yu Wang, Hua Zou, Xinzhu Yang and Hongkun Hao, from Bank and Cloud Payment, for our valuable discussions on business understanding and inspirations for the application design. The computing was executed on Everbright Data Haven (EDH), so we would like to express the deepest gratitude to the substantial help from EDH.


\bibliography{example_paper}
\bibliographystyle{icml2020}

\end{document}